# Generative Synthetic Augmentation using Edge-detected Semantic Labels for Segmentation Accuracy

Takato Yasuno[1]

[1] Yachiyo Engineering, Co., Ltd., RIIPS.

**Abstract.** In medical image diagnosis, pathology image analysis using semantic segmentation becomes important for efficient screening as a field of digital pathology. The spatial augmentation is ordinary used for semantic segmentation. Tumor images under malignant are rare and to annotate the labels of nuclei region takes much time-consuming. We require an effective use of dataset to maximize the segmentation accuracy. It is expected that some augmentation to transform generalized images influence the segmentation performance. We propose a "synthetic augmentation" using an edge feature detected semantic label-to-image translation, mapping from a semantic label with the edge structure to a real image. We demonstrate several segmentation algorithms applied to the nuclei semantic segmentation dataset that contains raw images and labels using synthetic augmentation in order to add their generalized images. We report that a proposed synthetic augmentation procedure is able to improve segmentation accuracy.

## 1 Introduction

### 1.1 Background and Related Work

**Semantic segmentation for digital pathology.** Microscopic pathology slides can capture the histologic details of tissues at high resolution. Owing to the rapid growing technology, whole slide imaging (WSI) becomes part of the ordinary procedure for clinical diagnosis of each disease. The automating image analysis accurately and efficiently remain significant opportunities. Recently, deep learning algorithms have shown great promise in pathology image analysis, such as in tumor region identification, metastasis detection, and patient prognosis. Many deep learning algorithms, including fully convolutional networks (FCN), U-Net, and newer inspired architectures, have been proposed to automatically segment pathology images. Among these algorithms, segmentation deep learning algorithms such as fully convolutional networks stand out for higher accuracy, computational efficiency, and end-to-end learnability. Thus, pathology image semantic segmentation have become a practical tool in WSI analysis.

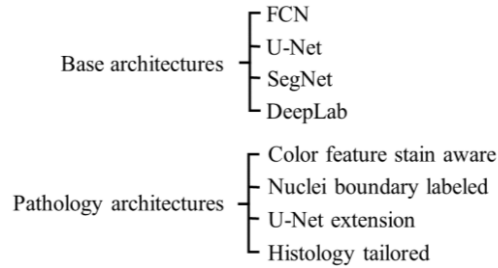

**Fig. 1.** Semantic segmentation architecture for digital pathology (tissue, nuclei, tumor).

Wang et al. reviewed as the applications of deep learning algorithms for pathology image segmentation process in WSI image analysis [1]. They pointed the image preprocessing that contains image normalization, shape augmentation, and color augmentation. There are several image shape augmentation methods. For example, a projective matrix transformation is possible such as scaling, translation, rotation, and affine transformations. The color augmentation is important to make the deep learning algorithm learn to adapt it, because pathology image may look very different due to different staining conditions and slide thickness. For example, hematoxylin and eosin (H&E) stained renal cell carcinoma pathology images are often classified into eosinophilic and basophilic subtypes, which are prone to be stained to by eosin (magenta) or hematoxylin (blue) , respectively, and thus have intrinsically different color distribution. There are several color augmentation methods such as adding a random mean, multiplying a random variation to each channel of each image, adding Gaussian noise, and so forth.

As shown in Figure1, there are useful architectures as semantic segmentation algorithms using deep learning for pathology image analysis. The first end-to-end and pixel-to-pixel semantic segmentation neural network is the Fully Convolutional Network (FCN) [2]. Many modifications have been made to FCN to further improve the segmentation performance. For example, U-Net [3] greatly increase the number of deconvolutional layers to propagate information to higher resolutions with convolution and max-pooling layers demonstrated in medical images. SegNet [4] refines the encoder-decoder network layers with skip connections, includes typically three layers: convolution, batch normalization, and ReLU. Recently, DeepLab replaced the deconvolutional layers with atrous (densely) convolution and atrous spatial pyramid pooling instead of standard convolution layers [5].

Inspired with these base network, there are many proposed deep neural network for pathology image segmentation frameworks such as color feature stain aware model [6][7], nuclei boundary labeled model [8][9], U-Net extension model [10], histology tailored model [11][12]. From a task point of view since 2017, Dabass et al. reviewed twenty pathological image segmentation studies using deep learning such as tissue, nuclei, glands, tumor, and various segmentation which applied to different pixel size input images and their accuracy measures such as Precision, Recall, F1-score, and Dice coefficient (Intersection of Union; abbreviate IoU) [13]. However, we have a common problem to be higher performance for the pathology image segmentation task so as to be accurate and stable process. More effective augmentation procedure is expected for computational pathology practitioners.

### 1.2 Synthetic dataset augmentation and conditional generative learning

In the Pascal Visual Object Classes (VOC) challenge dataset, Goyal et al. [14] studied that augmentation of the weakly annotated dataset with synthetic images minimizes both the annotation efforts and the cost of capturing images. Such a synthetic augmentation showed an increase in mean IoU from 52.8% to 55.4% by adding just 100 synthetic images per object class such as aeroplane, car, chair, dog, person and so forth. Unfortunately, the limitation of usability is to draw 3D objet models over a background with less reality and to learn the only the FCN-8s architecture.

Since 2014, there is 150 publication literatures on the generative adversarial network (GAN) in medical imaging by the review at January, 2019 [15], they categorized into synthesis, reconstruction, segmentation, classification and others. These high proportion of imaging modality contains MR, CT, histopathology, retinal fundus imaging and X-ray. Furthermore, there is a lot of studies about cross modality synthesis and conditional synthesis. For example, Tony et al. [16] proposed an automatic data augmentation method that uses GAN to learn augmentation that enable learn the available annotated samples efficiently. The architecture consists of a coarse-to-fine generator to capture the manifold of the training sets and generate generic augmented data. They showed the result on MRI image achieving improvements of 3.5% Dice coefficient compared to traditional augmentation method. But the main limitation is the usability under a customized architecture and the case of brain tumor segmentation. In pathology, various approaches have been proposed methods to virtually stain unstained specimens such as hematoxylin and eosin (H&E) [17], immunohistochemical staining (IHC) [18]. However, they have tended to focus on the speed and efficacy in diagnostic pathology workflow rather than the augmented accuracy improvement.

Using DCGAN for data augmentation, they achieved a significant improvement in classification accuracy compared to the baseline of standard data augmentation only [19]. They added synthetic data produced by their DCGAN, then the classification performance improved from around 80% to 85%, demonstrating the usefulness of GANs. However, they tended to focus on classification algorithms, it does not explain whether generative synthetic augmentation could improve performance using semantic segmentation algorithms. Such a segmentation accuracy have become important for diagnosis and prognosis. We often require to annotate the nuclei region of interest as a semantic label and take much workforces and long time.

In order to overcome the scarcity of tumor images captured on malignant, and the effective use of exist dataset, we have opportunities to generalize the feature of initial dataset using generative synthetic augmentation procedure. We propose a generative synthetic augmentation method in the semantic segmentation task for digital pathology, to clarify whether the generative synthetic augmentation can improve the segmentation performance. We demonstrate several base architectures to learn the both initial dataset and the generative synthetic images using L1-Conditional GAN (pix2pix).

## 2      Generative synthetic augmentation method

### 2.1    Proposed two generative synthetic augmentation using GAN

We think that approaches for generating a damage image include 1) reproducing the already acquired damage image (Similar augmentation), and 2) generating a future image degraded from the current damage grade (What-if degradation) and 3) what-if newer damage that does not yet exist (what-if newer). Here, 1) is close to data augmentation that has been performed as standard in supervised learning by rotation, X/Y translation, scaling, and so forth.

In case of 2), it is possible to simulate the situation where the deterioration has progressed several years ahead of the current state. Degraded state that has not yet been experienced, but generates an image of the state that has progressed one rank deterioration or the worst image when the management level is low, exceeds the scope where the supervised data exists. This is an attempt to eliminate any blind spots in the supervised learning. 3) is an approach that was not possible with supervised learning based on the experience data. Even in case of infrastructures that has not deteriorated, a new damaged image can be generated in order to prepare for future deterioration, and even if it has not yet been experienced, it enables to imagine a degraded future image. However, it is necessary to have a reality about where and how much tumor occurs in the stained images. This means that after acquiring images of potential tumor throughout histopathology. It is necessary to design a new possible tumor scenario and place the damage at the possible position. It is necessary to generate a new damaged image with ethics in order to make the pathologist uneasy about the tumor image without reality.

In this paper, we propose *two synthetic augmentation methods*. First, we propose to generate a synthetic tumor image dataset using the first replica method 1) of reproducing the current tumor images (we call "*replica synthetic augmentation*" that stands for *G0* below). Second, we propose to generate another hybrid augmentation method using both the standard shape augmented and the generative synthetic augmented procedure. Here, the shape augmentation stands for several well-known preprocesses such as rotation, X-axis reflection, upscale resize and random cropping (we call "*shape synthetic augmentation*" that denotes *G1* below).

### 2.2    Semantic segmentation architecture for demonstrations

In order to recognize the damage region of interest for social infrastructure, semantic segmentation algorithms are useful. We propose a synthetic augmentation method to generate fake images and labels using the L1-conditional GAN (pix2pix) to translate a label image with structure edge to a damaged image. We apply several existing per-pixel segmentation task based on transfer learning such as the FCN [2], U-Net [3], SegNet [4], and the dense convolution network such as the DeepLabv3+ResNet18, ResNet50 [5]. Among other candidate of backbones on the DeepLabv3+ architecture, as far as we try to compute the nuclei dataset, it could not improve their performance.

We compare the trained segmentation accuracy using initial dataset with the re-trained segmentation accuracy using synthetic augmentation added generated images.

We evaluate the both task performance to compute the similarity indexes between the ground truth damage region of interest (ROI) and the predicted region. Exactly, we compute the mean Intersection of Union (mIoU), class-IoU that consists the ROI and background. In order to analyze the property of synthetic augmentation, we compute the precision, recall and BF score. Therefore, using these existing segmentation algorithms, we get knowledge whether our method of synthetic augmentation can improve their segmentation accuracy or not.

### 2.3 Synthetic augmentation using semantic labels detected with structure edge

To train the DCGAN, we need more than 500 images and also they should have their stable angle. In the infrastructure deterioration process, a progressed damage is rare event and it is not easy to collect their damaged images more than even several hundreds. The eye-inspection view has various angle according to each field to monitor their social infrastructures. On the other hand, the image-to-image translation is possible for training a paired image dataset even with various inspection angle. This paper propose a synthetic augmentation method using L1-Conditional GAN (pix2pix). The original pix2pix paper translated form the input of edge images to shoe images [20]. And using the CamVid dataset, they translated from the semantic label to photo. However in case of damage images, we could not success such a naive translation. As shown Figure 2, this paper proposes the semantic label with structure edge as input of edge-detected semantic labels. This augmented label consists damage-ROI, enhanced structure edge, and background. We tried several edge detection method such as Gradient operators (Roberts, Prewitt, Sobel), Laplacian of Gaussian (LoG), Zero crossing, Canny edge and so forth [21]. This paper propose the Canny edge to extract the nuclear structure from tumor images. We combine the both semantic label and structure edge produced by the Canny edge detection into three class categorical label. We train the mapping from the semantic label with structure edge to tumor image.

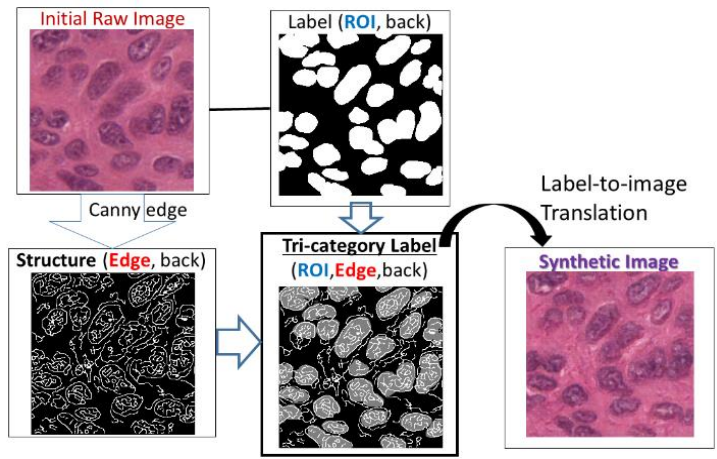

**Fig. 2.** Synthetic augmentation method using edge feature added semantic label-to-image translation on a nuclei in tumor stained tissue example.

As shown in Figure 2, we summarize a generative synthetic augmentation step as follows. First, we train one of semantic segmentation task using the initial dataset including with tumor images and semantic label. Second, we apply our replica and shape synthetic augmentation methods mapping to generate near-real images using L1-Conditional GAN from combined semantic label with structure edge. Third, we re-train another semantic segmentation task using the both initial dataset and our generated near-real tumor images dataset. The number of dataset is two or three times compared with the initial dataset, so as to extend an opportunity to learn the ROI and background feature between real tumor slide image and fake synthetic images.

## 3   Nuclei segmentation demonstration

### 3.1   Prepare edge-detected semantic labels and raw nuclei images

We use the one of open accessed dataset [9], which is the cancer tumor images and nuclei annotated labels. As shown in Figure 3, they are stained hematoxylin and eosin (H&E) imaged at 40x magnification. The dataset are partitioned with the 30 training images and the 14 test images. Kumer et al. [9] downloaded 30 whole slide images of digitized tissue samples of several organs from The Cancer Genomic Atlas (TCGA) [22]. These images came from 18 different hospitals. The training whole slide images contains four organ sub-types such as breast, liver, kidney, and prostate, while the test images includes three organ sub-types such as bladder, colon, stomach, these are different nuclei feature and background from the training data.

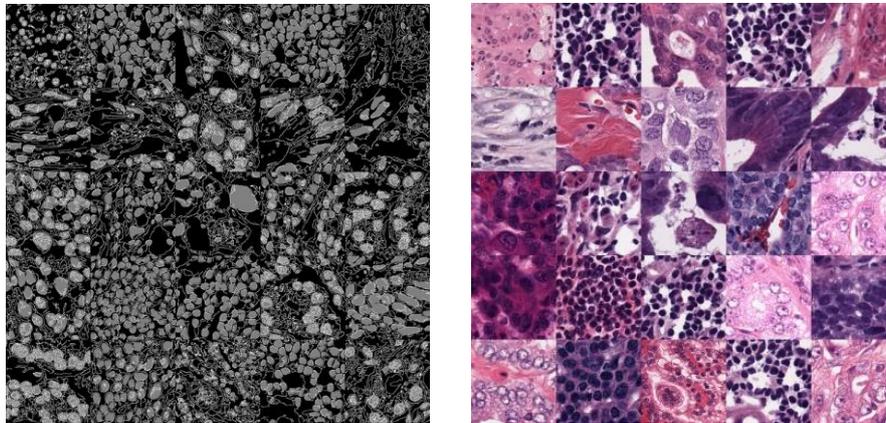

**Fig. 3.** Semantic labels (Left) with edge feature that contains the Canny edge structure (white color), the region of interest (ROI) of nuclei (silver color), and background (black color) in tumor.  Initial raw tumor images (Right) shuffle partially 25 selected from training dataset, where 30 raw images with size 1,000 by 1,000 pixels were extracted 480 block images sized 250 by 250 pixels. This dataset is opened at the website by Kumer, N., Verma, R. et al. [9].

Their dataset were annotated by Kumer et al. [9] into nuclei region of interest (ROI) and background two class labels over each raw tumor image for nuclei segmentation task. Each label has two category such as nuclei ROI (value 255), and background (value 0).The initial raw tumor image has whole image size of 1,000 by 1,000. In order to keep the pixel feature we prepare to extract 4 by 4 block images unified with size 250 by 250, so as to learn standard segmentation architecture to match the input size with 224 by 224. After these prepare, we have an initial training dataset with number of 480 images and labels. The total background pixel counts is 45,564 thousands, while the total nuclei pixel counts is 15,892 thousands. We compute two class weights that the nuclei-ROI is 1.933 and background is 0.674 divided by median pixel count.

### 3.2 Generate nuclei images using GAN

We applied the generative synthetic augmentation to the 480 tumor block image dataset based on the L1-Conditional GAN (pix2pix) [20]. We carried out label-to-mage translation from edge-detected semantic labels combined between the semantic nuclei-ROI and the structure edge by Canny edge operation into the real block tumor images. As showed in Figure 3 (left-side), the input semantic label has three classes that contains the structure edge (value 255), the nuclei-ROI (value 128), and the background (value 0), simultaneously. The output size is 256 by 256 by 3. We trained 200 epoch that took 19 hours. The L1 penalty coefficient is 100 at loss function.

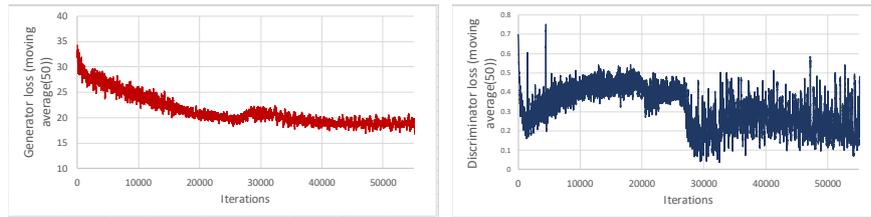

**Fig. 4.** Generator loss (Left) and Discriminator loss (Right) over training process of GAN

Figure 4 at the left-side shows the loss of generator over training process of L1-conditional GAN, where moving average 50 iterations to reduce their complexity of plot. After it repeated at around 55 thousands iterations, that is 115 epochs, the value of generator loss have reached at minimum and stable level. On the right-side, Figure 4 shows the loss of discriminator over training process of L1-conditional GAN. At the left half stage, the loss of discriminator is decreasing at 0.1 level so as to fool the weak synthetic tumor image, but after it repeated 30 thousands iterations, the discriminator loss value have approached real/fake trade-off movement so as to generate their elaborated synthetic images almost same as the real tumor images.

### 3.3 Accuracy comparison for generative output of test images

We predicted the test dataset of 30 slides, which is the four organ of the training dataset. We tried to predict these block test images with the number of 480, that equals 16 multiply 30, where we prepare to extract the 16 block images with 250 by 250 pixels from the 30 whole slide images with size 1,000 by 1,000 pixels. The weight between train and test is 95 : 5, so the number of test images is 24. The test images are shuffle sampling with fair variations.

Figure 5 shows the four raw tumor images and nuclei ROI labels sampled from the initial dataset, they are typical four organ sub-type within their training dataset.

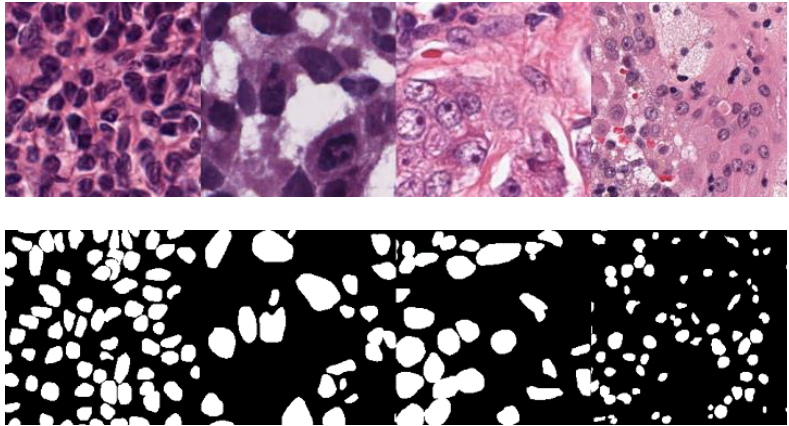

**Fig. 5.** Raw tumor images (upper four organ sub-type crops) and nuclei ROI semantic labels (lower four ground-truth), typically sampled from the initial test dataset.

**Table 1.** Test prediction from initial to augmented dataset, precision, recall, F1.

| architecture | augmented dataset | precision | | recall | | F1-score | | mean IoU | |
|---|---|---|---|---|---|---|---|---|---|
| | | ROI | backgrd | ROI | backgrd | ROI | backgrd | ROI | backgrd |
| FCN-8s | initial | 0.6178 | 0.7142 | 0.6711 | 0.7110 | 0.6411 | 0.7104 | 0.6128 | 0.8149 |
| | +replica(G0) | **0.6788** | **0.7632** | **0.7089** | **0.7452** | **0.6918** | **0.7526** | **0.6524** | **0.8325** |
| | +shape(G1) | **0.7475** | **0.8153** | **0.7569** | **0.7852** | **0.7515** | **0.7992** | **0.6912** | **0.8550** |
| U-Net | initial | 0.7221 | 0.7908 | 0.8260 | 0.8552 | 0.7647 | 0.8192 | 0.6703 | 0.8647 |
| | +replica(G0) | **0.8108** | **0.8700** | **0.8573** | **0.8682** | **0.8326** | **0.8688** | **0.7361** | **0.8835** |
| | +shape(G1) | **0.8455** | **0.8951** | **0.8721** | **0.8789** | **0.8581** | **0.8865** | **0.7639** | **0.8925** |
| SegNet-VGG16 | initial | 0.9950 | 0.9926 | 0.9808 | 0.9805 | 0.9877 | 0.9862 | 0.8669 | 0.9447 |
| | +replica(G0) | 0.9754 | 0.9858 | 0.9663 | 0.9675 | 0.9708 | 0.9765 | 0.8516 | 0.9398 |
| | +shape(G1) | **0.9997** | **0.9999** | **0.9932** | **0.9936** | **0.9964** | **0.9967** | **0.9209** | **0.9695** |
| DeepLabv3+ ResNet18 | initial | 0.9491 | 0.9679 | 0.9287 | 0.9309 | 0.9387 | 0.9489 | 0.8334 | 0.9303 |
| | +replica(G0) | **0.9695** | **0.9713** | 0.8335 | 0.8855 | 0.8929 | 0.9252 | 0.7932 | 0.9212 |
| | +shape(G1) | **0.9774** | **0.9864** | **0.9549** | **0.9554** | **0.9660** | **0.9705** | **0.8523** | **0.9387** |
| DeepLabv3+ ResNet50 | initial | 0.9945 | 0.9972 | 0.9701 | 0.9711 | 0.9821 | 0.9839 | 0.8889 | 0.9558 |
| | +replica(G0) | 0.9843 | 0.9805 | 0.8539 | 0.9064 | 0.9112 | 0.9406 | 0.8088 | 0.9292 |
| | +shape(G1) | **0.9980** | **0.9991** | **0.9814** | **0.9815** | **0.9896** | **0.9902** | **0.9042** | **0.9622** |

Table 1 shows the predicted results applied on the initial trained networks and another synthetic augmented trained network using our replica and shape synthetic augmentation methods. Viewing on the FCN-8s, SegNet-VGG16, and DeepLabv3+ based on ResNet18 and ResNet50, our replica and shape synthetic augmentation methods can perform higher-precision index value from the viewpoint of the nuclei-

ROI and background. Moreover, the FCN-8s and SegNet-VGG16 indicate the higher recall index value. These lead to the higher F1-score and mean IoU. Thus, we promised that the shape synthetic augmentation method toward the test dataset is effective for the base architectures.

### 3.4 Generative output images of test dataset

Figure 6 shows the overlay of two labels between the ground truth region of nuclei interest and the predicted region by the U-Net trained segmentation networks. The white region is good prediction to match between the ground truth and the nuclei prediction. In contrast, the green region is over precision and the magenta region is less recall. From top to bottom, the first output stands for the initial dataset based prediction, next, the middle result denotes our replica synthetic augmented prediction. At the bottom, it indicates our shape synthetic augmented prediction. Compared these predicted results, using our shape synthetic augmentation method, their extra-predicted green region and less-recall region are decreasing gradually.

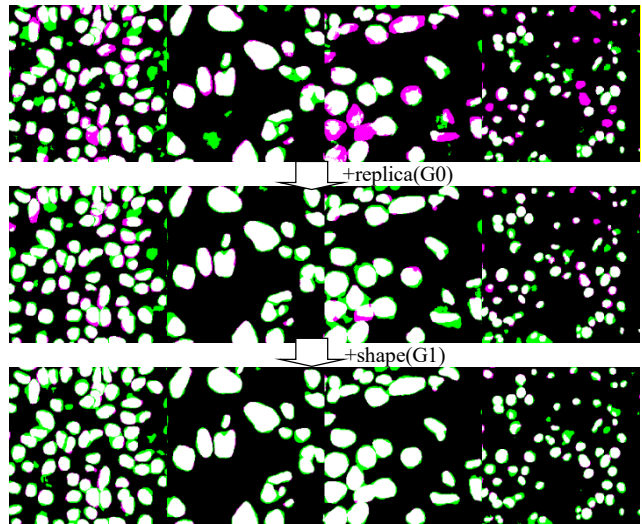

**Fig. 6.** Overlay between ground truth and U-Net predicted mask, the initial segmentation (top) with our replica augmentation (middle), and shape synthetic augmentation (bottom).

In the same manner, Figure 7 shows the nuclei predicted results by the trained SegNet-VGG16. Furthermore, Figure 8 shows the predicted results by the trained DeepLabv3+ResNet50. Therefore, the shape synthetic augmentation methods can always reduce the green region (over-precision) and the magenta region (less-recall), so it enables to maximize the white region (accurate prediction to match the ground truth) on these base architectures of semantic segmentation algorithms.

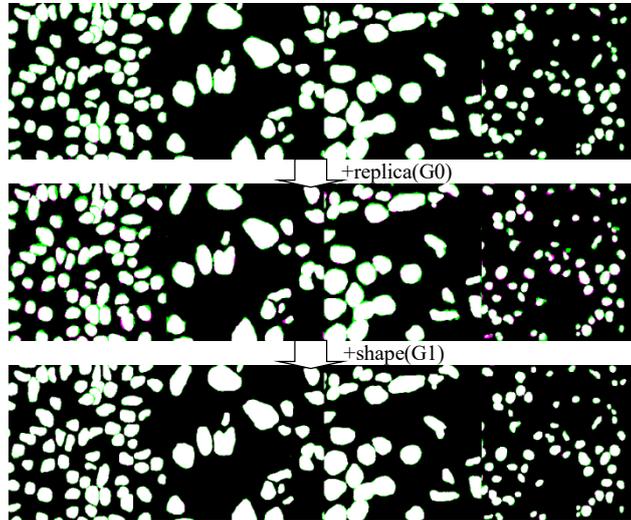

**Fig. 7.** Overlay between ground truth and SegNet-VGG16 predicted mask, the initial (top) with our replica augmentation (middle), and our shape synthetic augmentation (bottom).

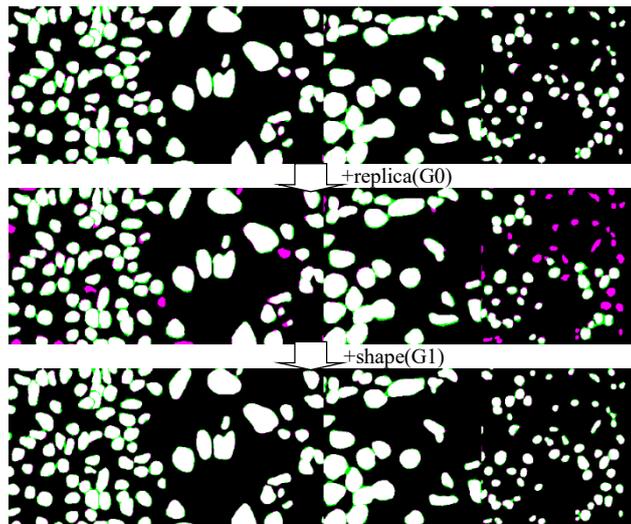

**Fig. 8.** Overlay between ground truth and DeepLabv3+ with the backbone of ResNet50 predicted mask, to compare the initial segmentation (top) with our replica synthetic augmentation (middle), and our shape synthetic augmentation (bottom).

## 4 Conclusion

### 4.1 Concluding Remarks

The author proposes a synthetic augmentation procedure using the input label revised L1-Conditional GAN algorithm, which maps edge-detected semantic labels to real tumor images. Our proposed labels consist of three classes such as the region of interest (ROI), structure edge feature, and the background. We propose a Canny edge to extract the feature of structure edge from photos such as stain tissue images. We demonstrated a synthetic augmentation procedure using L1-Conditional GAN applied to datasets for semantic segmentation. We demonstrated that predicted results using our the shape synthetic augmentation methods are possible to improve the segmentation accuracy over several base architectures, such as FCN-8s, U-Net, SegNet-VGG16, and DeepLabv3+ResNet50. From a performance merit of view, our synthetic augmentation methods can reduce the green region (over-precision), and hence it improves the precision accuracy of several segmentation algorithms.

### 4.2 Future Works

We aim to tackle the development of a more general-purpose diagnosis application for digital pathology. This paper focused on the nuclei segmentation dataset in tumor images at the stained tissue. We applied our generative synthetic augmentation methods to the basic five architectures, instead more general feasibility study remains to be conducted toward recent refined pathology architectures. There are opportunities to apply other nuclei segmentation of malignant on different disease slides images. Furthermore, another what-if type augmentation study remains, for example, malignant scenario augmentation what if new tumor occurred on benign images.

## References

1. Wang, S., Yang, D.M., Rong, R. et al., Pathology Image Analysis Using Segmentation Deep Learning Algorithms, American Journal of Pathology, Vol.189, No.9, (2019).
2. Shelhamer, E., Long, J., Darrell, T., Fully Convolutional Networks for Semantic Segmentation, IEEE Trans Pattern Anal Mach Intell, 39, pp640-651, (2017).
3. Ronneberger, O., Fischer, P., Brox, T., U-Net: Convolutional Networks for Biomedical Image Segmentation, edited by Navab, N. et al. In Medical Image Computing and Computer-Assisted Intervention (MICCAI) 2015, Springer, pp234-241, (2015).
4. Badrinarayanan, V., Kendall, A., Cipolla, R., SegNet: A Deep Convolutional Encoder-decoder Architecture for Image Segmentation, IEEE Trans Pattern Anal Mach Intell, 39, pp2481-2495, (2017).
5. Chen, L.C., Papandreou, G., Kokkinos, I. et al., DeepLab: Semantic Image Segmentation with Deep Convolutional Nets, Atrous Convolution, and Fully Connected CRFs, IEEE Trans Pattern Anal Mach Intell, 40, pp834-848, (2018).
6. Lahiani, A., Gildenblat, J., Klaman, I. et al., Generalizing Muli-stain Immunohistochemistry Tissue Segmentation using On-shot Color Deconvolution Deep Neural Networks, (2018), arXiv:1805.06958v3.
7. Graham, S., Rajpoot, N., SAMS-NET: Stain-aware Multi-scale Network for Instance-based Nuclei Segmentation in Histology Images, April, (2018), DOI:10.1109/ISBI. 2018.8363645.
8. Cui, Y., Zhang, G., Liu, Z. et al., A Deep Learning Algorithm for One-step Countour Aware Nuclei Segmentation of Histopahological Images, (2018), arXiv:1803.02786v1.

9. Kumar, N., Verma, R., Sharma, S. et al., A Dataset and a Technique for Generalized Nuclei Segmentation for Computational Pathology, IEEE Transaction on Medical Imaging, Vol.36, No.7, (2017).
10. Alom, M.Z., Aspiras, T., Taha, T.M. et al., Advanced Deep Convolutional Neural Network Approaches for Digital Pathology Image Analysis: A Comprehensive Evaluation with Different Use Cases.
11. Chan, L., Hosseini, M.S., Rowsell, C. et al., HistSegNet: Semantic Segmentation of Histological Tissue Type in Whole Slide Images, pp10662-10671, ICCV2019, (2019).
12. Graham, S., Chen H., Gamper, J. et al., MILD-Net: Minimal Information Loss Dilated Network for Gland Instance Segmentation in Colon Histology Images, Medical Image Analysis, (2018).
13. Dabass, M., Vashisth, S., Review of Histopathological Image Segmentation via current Deep Learning Approaches, May (2019): DOI:10.1109/CCAA.2018.8777616.
14. Goyal M, Rajpura P, Bojinov H et al. : Dataset Augmentation with Synthetic Images Improves Semantic Segmentation, arXiv:1709.00849v3, (2018).
15. Yi X, Walia E, Babyn : Generative Adversarial Network in Medical Imaging: A Review, arXiv:1809.07294v4, (2019).
16. Mok T, Chung A : Learning Data Augmentation for Brain Tumor Segmentation with Coarse-to-Fine Generative Adversarial Networks, arXiv:1805.11291v2, (2018).
17. Bayramoglu N, Kaakinen M, Eklund L et al. : Towards Virtual H&E Staining of Hyper-spectral Lung Histology Images Using Conditional Generative Adversarial Networks, ICCV, (2017).
18. Seneras C, Niazi M.K.K., Sahiner B et al. : Optimized Generation of High-Resolution Phantom Images Using cGAN: Application to Quantification of Ki67 Brest Cancer Images, Plos One, (2018).
19. Frid-Adar, M., Diamant, I. et al, GAN-based Synthetic Medical Image Augmentation for increased CNN Performance in Lesion Classification, CVPR, (2018).
20. Isola, P., Zhu, J-Y. et al., Image-to-image Translation with Conditional Adversarial Network, CVPR, (2017).
21. Gonzalez, R.C., Woods, R.E. : Digital Image Processing, 4th Global Edition, Pearson, (2018).
22. The Cancer Genome Atlas, accessed on March 26, (2020). https://www.cancer.gov/about-nci/organization/ccg/research/structural-genomics/tcga.
23. Zhou, S.K., Greenspan, H., Shen, D., Deep Learning for Medical Image Analysis, The Elsevier and MICCAI Society Book Series, Academic Press, (2017).
24. Kumer, V., Abbas, A.K., Aster, J.C., Robbins Basic Pathology 10th edition, Elsevier, (2018).
25. Foster, D., Generative Deep Learning: Teaching Machines to Paint, Write, Compose and Play, O'Reilly, (2019).
26. Holzinger A, Goebel R, Mengel M, Muller H (Eds.) : Artificial Intelligence and Machine Learning for Digital Pathology: State-of-Art and Future Challenges, Lecture Notes in Computer Science 12090, Springer, (2020).